\documentclass[conference]{IEEEtran}
\IEEEoverridecommandlockouts
\usepackage{cite}
\usepackage{amsmath,amssymb,amsfonts}
\usepackage{multirow}
\usepackage{algorithmicx}
\usepackage{graphicx}
\usepackage{textcomp}
\usepackage{algorithm}
\usepackage{latexsym}
\usepackage{algpseudocode}
\usepackage{xcolor}

\def\BibTeX{{\rm B\kern-.05em{\sc i\kern-.025em b}\kern-.08em
    T\kern-.1667em\lower.7ex\hbox{E}\kern-.125emX}}
\begin{document}

\title{Graph Convolutional Networks in Feature Space for Image Deblurring and Super-resolution\\
}

\author{\IEEEauthorblockN{Boyan Xu}
\IEEEauthorblockA{\textit{Department of Electrical and Electronic Engineering} \\
\textit{The University of Manchester}\\
Manchester, UK \\
boyan.xu@postgrad.manchester.ac.uk}
\and
\IEEEauthorblockN{Hujun Yin}
\IEEEauthorblockA{\textit{Department of Electrical and Electronic Engineering} \\
\textit{The University of Manchester}\\
Manchester, UK \\
hujun.yin@manchester.ac.uk}
}

\maketitle

\begin{abstract}
Graph convolutional networks (GCNs) have achieved great success in dealing with data of non-Euclidean structures. Their success directly attributes to  fitting graph structures effectively to data such as in social media and knowledge databases. For image processing applications, the use of graph structures and GCNs have not been fully explored. In this paper, we propose a novel encoder-decoder network with added graph convolutions by converting feature maps to vertexes of a pre-generated graph to synthetically construct graph-structured data. By doing this, we inexplicitly apply graph Laplacian regularization to the feature maps, making them more structured. The experiments show that it significantly boosts performance for image restoration tasks, including deblurring and super-resolution. We believe it opens up opportunities for GCN-based approaches in more applications. 
\end{abstract}


\section{Introduction}
Graph convolutional networks (GCNs) have recently been shown outstanding ability in dealing with data of non-Euclidean structures, such as point clouds and graphs. Recently they have gained much increased attention in the signal/image processing and machine learning communities. Many existing applications of GCNs have focused on graph data or data exhibiting graph structures, such as social networks \cite{hamilton2017inductive, kipf2016semi}, physical systems \cite{sanchez2018graph}, and knowledge graphs \cite{hamaguchi2017knowledge}.

\begin{figure}
\centerline{\includegraphics[width=3in]{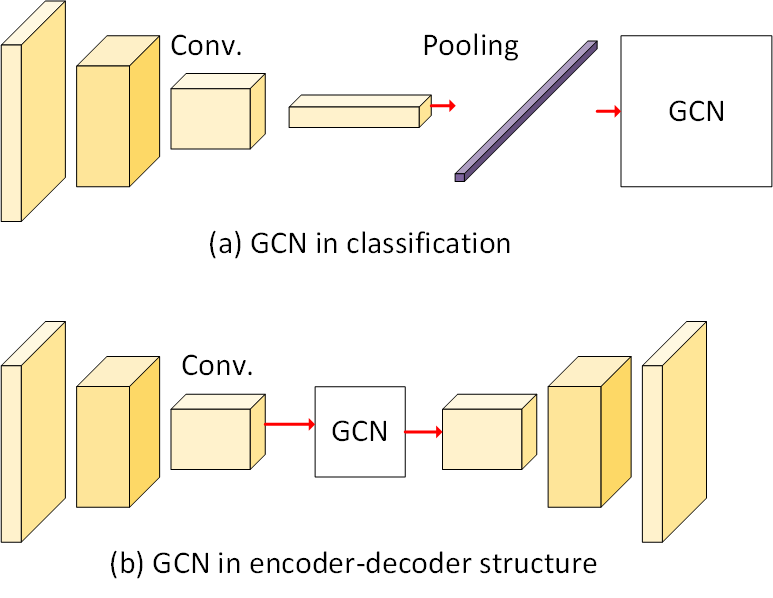}}
\caption{Comparison of GCN in classification and the encoder-decoder structure. Since the data processed by pooling and fully connected network (FCN) can be processed by GCN, we can infer that the latent information in the encoder-decoder can be better represented by GCN.}
\label{fig:head}
\end{figure}

The main advantage of graph networks is to express dissemination among information and interaction of data; hence GCNs are powerful tools to represent the intra-relationship of input data. In \cite{lee2018multi}, GCN was used in classification to describe  relationships between multiple labels. In this case, the original data was not graph-structured (i.e., images in the Euclidean space), while high-level features were abstracted and further processed by knowledge graphs. Similar work can be also seen in \cite{wang2018zero}. GCN in image classification is often applied on the result of encoder, as indicated in Fig. \ref{fig:head} (a), and the existing research suggests that GCN works well on these encoded data. Such applications are based on high-level semantics. By comparing network structures of classification and image restoration, we can reasonably argue that semantic relationships should also exist in low-level features, for instance, intermediate feature maps in the convolutional neural networks (CNNs). In this paper, we explore the use of GCNs in an encoder-decoder structure, as illustrated in Fig. \ref{fig:head} (b). 

In the forward process of a CNN, features produced by the convolutions may be used to infer their topological relationships. Some features may be more important and subsequent layers may depend more on these key features. These topological relationships can be described by graph networks. Therefore, we propose a GCN-based encoder-decoder network to exploit such relationships among features. For efficiency, we first produce an artificially constructed graph structure and then fit features into the graph, followed by their corresponding weight updating. By doing this, the features extracted contain certain structural relationships useful to many image restoration tasks such as deblurring and super-resolution. To our best knowledge, there is no similar approach before. Such use of GCNs also broadens the application of GCNs.

\begin{figure*}
\centerline{\includegraphics[width=\textwidth]{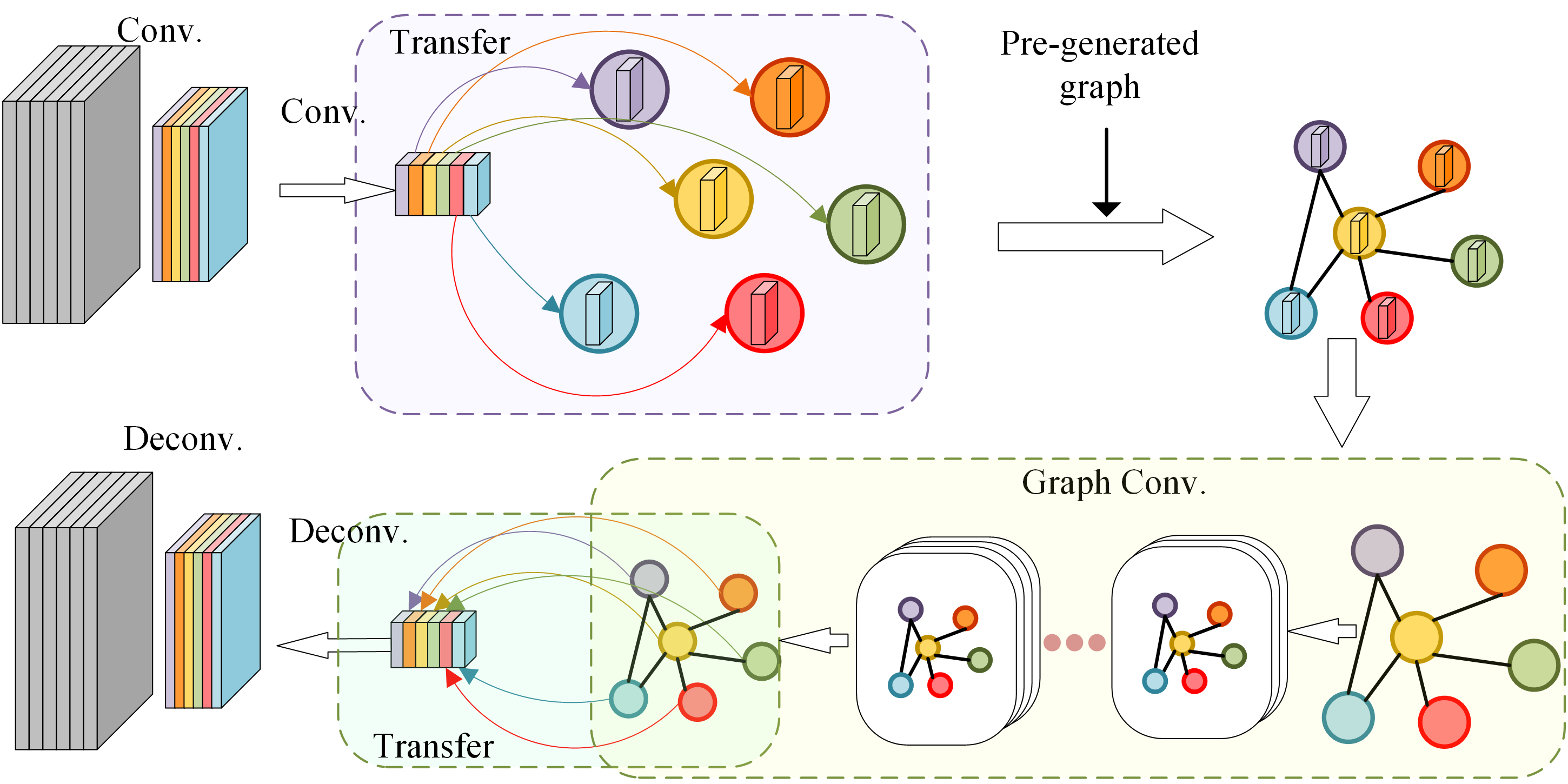}}
\caption{Proposed method converts feature maps from CNN into a graph, treating each channel as an identity or node and connecting them by a pre-defined adjacency matrix. The output layer of GCN is converted back to feature maps.}
\label{fig:how2convert}
\end{figure*}

The proposed network adds graph convolutions by converting feature maps to vertexes of a pre-generated graph to extract topological structures of the features. By doing this, we inexplicitly apply graph Laplacian regularization to the feature maps, making them more structured. Furthermore, we use residual learning to moderately deepen the graph network in order to increase performance. The experiments show that GCN in the feature space can significantly improve the performance in the task of image restoration (we use image deblurring and super-resolution as examples). We also analysed the relationship between our method and the channel attention structure.

We summarise our contributions as follows:
\begin{itemize}
\item We proposed a concept to transfer feature maps into vertexes of a pre-generated graph for graph convolution, in order to extract inter-relations of the features.
\item We proposed a new framework to fit graph network into the encoder-decoder structure, inexplicitly applying graph Laplacian regularization to the feature maps. This can be also regarded as an expansion of channel attention.
\item We applied GCN-enhanced deep network to image restoration tasks, in particular deblurring and super-resolution, and extensive experiments demonstrated the superiority of the proposed network compared to the state-of-the-art methods.
\end{itemize}

\section{Related Work}
\subsection{Graph Convolutional Networks (GCNs)}
From the perspective of aggregator, GCNs can be divided into spectral-based and spatial-based. Authors in \cite{bruna2013spectral} firstly developed graph convolutions based on spectral graph theory using the Fourier basis of a given graph in the spectral domain. Many extensions subsequently apply extensions, improvements and approximations on spectral-based GCNs \cite{kipf2016semi, henaff2015deep}. Spatial-based GCNs \cite{hamilton2017inductive, gao2018large} directly define graph convolution operations on the graph by operating on spatially close neighbours.

\begin{figure*}
\centerline{\includegraphics[width=\textwidth]{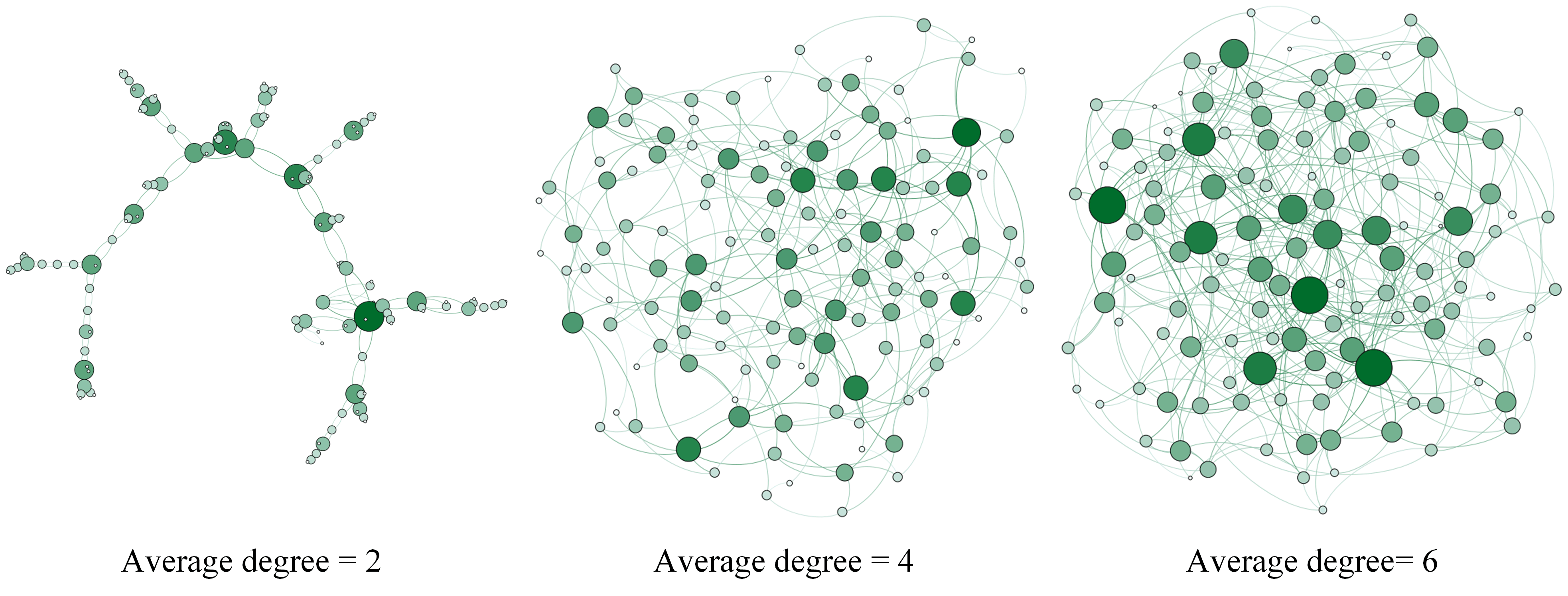}}
\caption{Our adopted graph with different mean node degrees, which are 2, 4, and 6 respectively. The mean of degree cannot be odd number \cite{watts1998collective}. Graph with mean degree more than 6 has very densely connected links, which is not easy to illustrate. The size of nodes is linear with the degree of the graph.}
\label{fig:graph}
\end{figure*}

In the recent rapid and fruitful development of GCNs, most methods employed shallow GCNs. Some attempted different ways of training deeper GCNs \cite{hamilton2017inductive,xu2019graph}. However, these networks are limited to 10 layers in depth before  performance degrades. Inspired by the benefit of training deep CNN-based networks \cite{he2016deep}, DeepGCNs \cite{li2019deepgcns} proposed to train a very deep GCN (56 layers) by adapting residual/dense connections (ResGCN/DenseGCN) to GCN. 

\subsection{Image Deblurring}
Image deblurring is a challenging task due to its ill-posed nature. It aims to recover the latent sharp image from a degraded input due to factors such as camera or object motion, which hinders many computer vision tasks including object detection and classification. To restraint the solution space, many prior based methods have been proposed, for instance, dark channel prior \cite{pan2016blind} and extreme channel prior \cite{cai2019extreme}. With the advent of deep neural networks (DNNs) \cite{krizhevsky2012imagenet, lecun2015deep}, some methods use deep neural networks to help find blur kernels and restore sharp images \cite{sun2013edge, li2019blind}. In \cite{nah2017deep}, end-to-end deblur CNN \cite{lecun1998gradient} was proposed, directly recovering sharp image without considering blur kernels. Kupyn \emph{et al.} \cite{kupyn2018deblurgan, kupyn2019deblurgan} adapted generated adversarial network (GAN) in image deblurring. In \cite{zhang2018dynamic}, recurrent neural networks were adopted, inspired by infinite impulse response (IIR). Tao \emph{et al.} \cite{tao2018scale} proposed scale-recurrent network with parameter sharing, and Gao \emph{et al.} \cite{gao2019dynamic} further adopted densely connected networks in this framework. 

\subsection{Image Super-Resolution}
Image super-resolution (SR), which refers to the process of estimating a high-resolution (HR) image from its low-resolution (LR) input, has attracted extensive attention in the computer vision community due to its wide range of applications. To restore realistic HR details, early approaches rely on interpolation techniques based on sampling theory \cite{zhang2006edge}. Natural image statistics was adopted in \cite{tai2010super} to reconstruct better high-resolution images. However, these methods have limitations in predicting realistic and reasonable textures due to the large solution space. Based on the success of CNNs, many CNN-based learning methods have been developed for the SR task. Dong \emph{et al.} proposed SRCNN \cite{dong2015image} to adopt deep convolutional network into solving image super-resolution. In \cite{kim2016accurate}, Kim \emph{et al.} observed that increasing network depth showed a significant improvement in accuracy, and further proposed VDSR. Haris \emph{et al.} \cite{haris2018deep} proposed a deep back-projection method by using iterative up- and down-sampling. Lim \emph{et al.} \cite{lim2017enhanced} proposed EDSR, which improved the performance by removing unnecessary modules in the conventional residual networks.


\begin{figure*}
\centerline{\includegraphics[width=\textwidth]{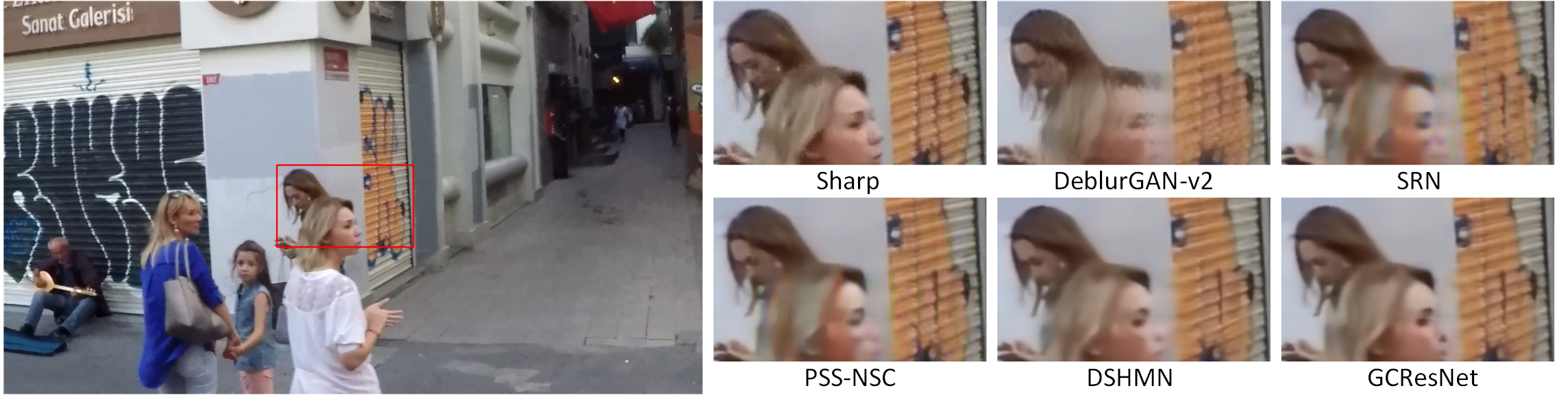}}
\caption{Visual comparison of image deblurring on GoPro, with DeblurGAN-v2 \cite{kupyn2019deblurgan}, SRN \cite{tao2018scale}, PSS-NSC \cite{gao2019dynamic}, and DSHMN \cite{zhang2019deep}. Our method produced clearer details especially on person's hair and stripes on the shutter door.}
\label{fig:deblurres1}
\end{figure*}

\section{Graph Network on Feature Maps}
The concept of our entire network is shown in Fig. \ref{fig:how2convert}. Convolutional layers produce high dimensional feature maps. Each feature map is transferred into a set of independent vertexes, and features are then connected via a pre-generated graph. By doing so, the features extracted by the convolutional layer become a structured graph network, which can be processed by graph convolutions. After several graph convolutions, we transfer the data from nodes back to feature maps in the same order. These features are further processed by decoder network and finally reconstructed in the output image for deblurring or super-resolution. 

\subsection{Conversion between feature map and graph nodes}
Since we use graph convolutions on features of a CNN, one of the difficulties is the conversion between feature maps and graph nodes, i.e., basic units of a graph. Usually, the implementation of graph convolution is based on adjacency matrix and degree matrix of a given graph, thus the data structure in training a deblurring or SR network need to fit in those matrix operations. In addition, using too many filters to convolve can lead to much increased dimensions and computational complexity. For instance, using 3 filters to convolve a 2D grey-scale image can produce a 3-dimensional matrix. To deal with these problems, we propose a new method, given in Algorithm \ref{alg::convert}. Consider an input $X$, by converting dimension $C$ to the graph, yielding a new dimension $F$, the data can appear in $C \times F$ structure with each element a feature map for left multiplication, and is the same as the mathematical form given in \cite{kipf2016semi}.

\renewcommand{\algorithmicrequire}{\textbf{Input}}
\renewcommand{\algorithmicensure}{\textbf{Output}}
\begin{algorithm}[h]
  \caption{Conversion between feature maps and nodes} 
  \label{alg::convert}
  \begin{algorithmic}
    \Require
      $X \in \mathbb{R}^{B_s \times C \times W \times H}$: feature map, where $B_s$ is batch size, $W$ and $H$ is width and height of input feature, $C$ is the number of features;
    \Ensure
      $X^{out} \in \mathbb{R}^{B_s \times C \times W \times H}$: the feature map after graph convolution.
    \State Convert $X \in \mathbb{R}^{B_s \times C \times W \times H}$ to $X^{*} \in \mathbb{R}^{B_s \times H \times W \times C}$;
    \For{all data in an epoch}
      \State Convert $X$ to $X^{*} \in \mathbb{R}^{B_s \times H \times W \times C \times F}$, added $F$ as graph feature dimension;
      \For{numbers of GC layers}
        \State $X^{*} \leftarrow GraphConv(X^{*})$;
      \EndFor
      \State Convert $X^{*}$ to $X \in \mathbb{R}^{B_s \times H \times W \times C}$, with graph feature dimension $F$ removed;
    \EndFor
    \State Convert $X \in \mathbb{R}^{B_s \times H \times W \times C}$ to $X \in \mathbb{R}^{B_s \times C \times W \times H}$;
  \end{algorithmic}
\end{algorithm}
\subsection{Graph Convolution}
\subsubsection{Pre-generated Graph}
Graph structure is important in GCNs. However, we find that using any suitable structure from natural laws and adding graph convolutions can help improve the CNN performance in image restoration. Generating the graph structures at each iteration during training can be extremely computationally intensive and may not produce stable results. We propose to use a pre-generated graph structure throughout to alleviate the burden of generating graphs dynamically. With a pre-generated graph, transforming conventional convolutions to graph convolutions incurs little extra cost. Here, we use the Watts–Strogatz (WS) model \cite{watts1998collective}, a random generated graph that has small-world network properties, such as clustering and short average path length. For a small-world graph, the average minimum path length is usually small and produces some hub nodes, to reflect importance of the features. We generate the graph based on number of features. For instance, for a 96-channel feature map, the graph has 96 nodes. Since we randomly create the graph for training, we have the following theorem to show that different random graphs would not result in marked differences in graph properties, determined by degree centrality \cite{borgatti2005centrality}, which is defined as the number of links upon a node.

\newtheorem{theorem}{Theorem}[section]
\begin{theorem}
Assume $x_1, x_2, x_3, ... x_n$ are the degrees of a graph of $n$ nodes, in Watts-Strogatz model, and $M^i$ is the average degree of the $i^{th}$ random graph, $M^i = \frac{1}{n} \sum_{j=1}^n x_j^i$. With large or increasing value of $n$, $M^i \to k$, a constant. The degree of graphs conforms to Gaussian distribution.
\end{theorem}

Theorem 3.1 can be easily proved by the \emph{Law of Large Numbers}. 

The core properties of a graph is determined by the degree of the graph. The average degree of a set of random graphs of same number of nodes becomes stable when the number increases. There are some nodes that are more significant than others, thus the weights of CNN will adapt to the graph structure when training with a new graph. The order of nodes does not affect much the network performance. Exemplar graphs of different average degrees are illustrated in Fig. \ref{fig:graph}.

\subsubsection{Aggregator and Updator}
For GCN, propagation contains aggregators to obtain hidden states of nodes. Various GCNs utilise different aggregators to gather information from each node’s neighbours and specific updaters to update nodes’ weights. Kipf \emph{et al.} \cite{kipf2016semi} developed an aggregator for spectral GCNs. Consider an undirected graph $\mathcal{G}$, our aggregator is given as
\begin{equation}
T = \tilde{D}^{-\frac{1}{2}} \tilde{A} \tilde{D}^{-\frac{1}{2}} X.
\end{equation}
where $\tilde{A} = A + I_N$ denotes the adjacency matrix of the graph $\mathcal{G}$ with added self-connections produced by identity matrix $I_N$. $N$ is the number of nodes. $T$ is the aggregator. Based on the renormalisation trick proposed in \cite{kipf2016semi}, $I_N + D^{-\frac{1}{2}} A D^{-\frac{1}{2}}  \to \tilde{D}^{-\frac{1}{2}} \tilde{A} \tilde{D}^{-\frac{1}{2}}$, where $D$ is the degree matrix. From \cite{kipf2016semi, zhou2018graph}
\begin{equation}
X^{l+1} = T^l \Theta^l.
\label{eq:H}
\end{equation}
where $\Theta^l \in \mathbb{R}^{C\times F}$ is a matrix of graph convolution (GC) filter parameters of $C$ input channels and $F$ filters in the $l^{th}$ GC layer, $T^l$ is the aggregator of the $l^{th}$ layer, $X^{l+1}$ is the convolved matrix after the $l^{th}$ layer. Thus we get
\begin{equation}
GraphConv(X) = \tilde{D}^{-\frac{1}{2}} \tilde{A} \tilde{D}^{-\frac{1}{2}} X \Theta.
\label{eq:GC}
\end{equation}
as a representation of the graph convolution used in the proposed network. 

\begin{figure*}
\centerline{\includegraphics[width=\textwidth]{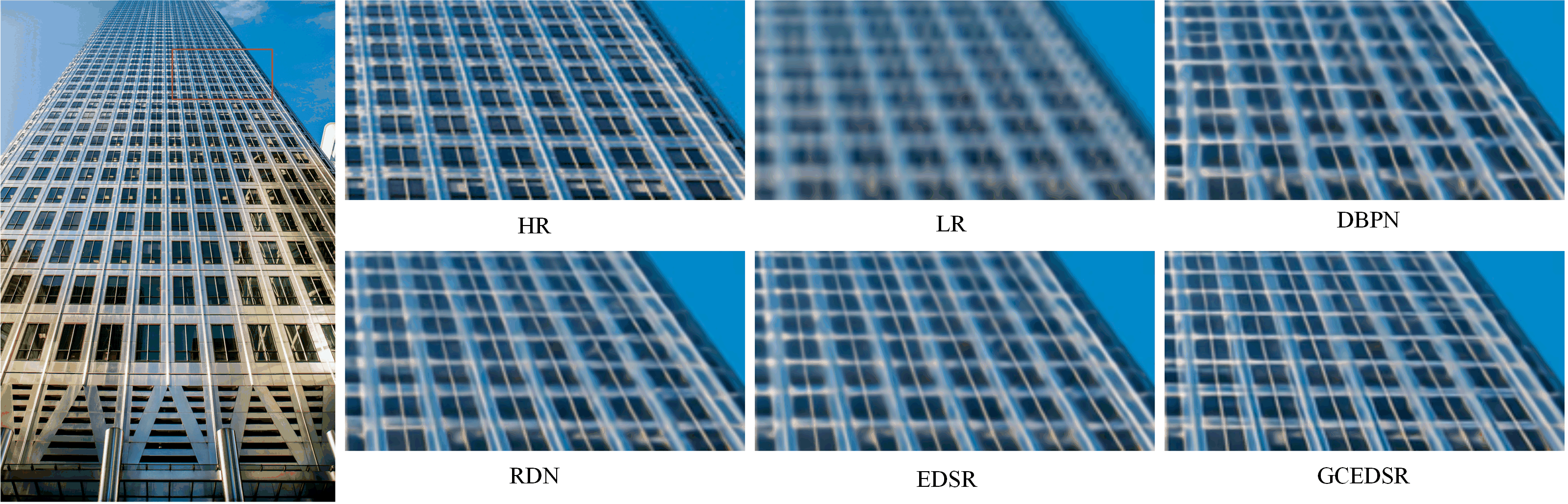}}
\caption{Visual comparison of image super-resolution on Urban100, with RDN \cite{zhang2018residual}, DBPN \cite{haris2018deep}, and EDSR \cite{lim2017enhanced}.}
\label{fig:srres}
\end{figure*}

\subsubsection{Deep GCN}
To further improve the performance, we also consider ResGCN \cite{li2019deepgcns} to make the network deeper. As being analysed in \cite{li2019deepgcns,li2020deepergcn}, deepening the network is useful. We believe that such conclusion is also applied in GCNs. Compared with \cite{li2019deepgcns}, our network faces to a fresh challenge with pre-generated graph, and the aggregator will be different from the original ResGCN. Thus, we removed normalization and limited the number of ResGCN blocks within 10 to avoid computational complexity. Based on Eq. \ref{eq:GC}, the ResGCN block used in this paper can be given as 
\begin{equation}
X^{out} = GraphConv(\alpha(GraphConv(X^{in}))) + X^{in}.
\end{equation}
where $\alpha(\cdot)$ denotes the activation function, e.g. ReLU. In this paper, we compared and analysed the influence of different number of ResGCN blocks in the experiments.

\section{Experiments}
\subsection{Experiments on Deblurring}
\subsubsection{GCResNet}
\begin{figure}
\centerline{\includegraphics[width=3.5in]{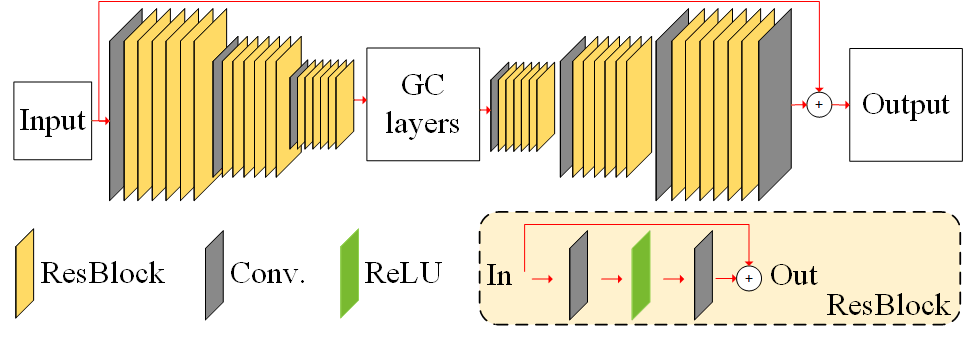}}
\caption{Structure of proposed network for image deblurring. The red lines denote skip connections.}
\label{fig:netdeb}
\end{figure}

We adopted graph convolutions in the residual blocks (ResBlocks) in our network for deblurring and term it as Graph Convolution ResNet (GCResNet). We removed normalization layers based on the analysis of \cite{tao2018scale, lim2017enhanced}. The network is based on an encoder-decoder structure with residual link from the import of the network to the last convolution layer. We used 18 ResBlocks in encoder and 18 ResBlocks in decoder. We adopted graph convolution layers between encoder and decoder. The network structure is shown in Fig. \ref{fig:netdeb}.  We used MSE loss, as it is the most suitable loss function and widely used in image deblurring \cite{gao2019dynamic, tao2018scale}. 

We evaluated the GCResNet on the GoPro dataset \cite{nah2017deep}, the most used end-to-end deblurring dataset. There are 2,103 pairs of sharp and blur images for training and 1,111 pairs for evaluation. The set was collected by averaging sharp images from videos thus is more realistic compared to other synthetic blurring datasets \cite{schuler2015learning}. We further tested the model on the Human-aware Image Deblurring (HIDE) dataset \cite{shen2019human}, which covers both wide-range and close-range scenes. The HIDE dataset has two parts: HIDE I (1304 long-shot pictures) and HIDE II (7118 close-ups pictures). We combined HIDE I and HIDE II, leading to 6397 images for training, 1063 of HIDE I and 962 of HIDE II for testing.

\subsubsection{Training Details}
Implementation of the proposed method consists of production of a graph network and training the network. We used MATLAB \cite{higham2016matlab} to produce the WS graph with $\rho = 0.9$ (slightly different values would not result in significant differences). 
We implemented the proposed network by Pytorch on a NVIDIA Tesla P100 GPU. During training, we randomly cropped a $256 \times 256$ region from a blurred image and used it and its ground truth image at the same location as the training input. The batch size was set to 12. All weights were initialized by the Xavier method \cite{glorot2010understanding}, and biases were initialized to zero. The network was optimized by using the Adam \cite{kingma2014adam} with default setting $\beta_1=0.9$, $\beta_2=0.999$ and $\epsilon=10^{-8}$. The learning rate was initially set to 0.0001 and linearly decayed to 0. 
\subsubsection{Results}
We used the Peak Signal-to-Noise Ratio (PSNR), Structural SIMilarity (SSIM) and Feature SIMilarity (FSIM) index \cite{zhang2011fsim} for image quality assessment. The proposed network was compared with the mainstream methods: DeepDeblur \cite{nah2017deep}, SRN-Deblur \cite{tao2018scale}, PSS-NSC \cite{gao2019dynamic}, DeblurGANv2\cite{kupyn2019deblurgan} and SVRNN \cite{zhang2018dynamic}. Results are shown in Table. \ref{result gopro}. The proposed method has the best performance. A visual comparison is shown in Fig. \ref{fig:deblurres1}, indicating that GCResNet has restored clearer and sharper details.
\begin{table*}
\renewcommand{\arraystretch}{1.3}
\caption{Testing results on GoPro dataset}
\label{result gopro}
\centering
\begin{tabular}{c|ccccccc}
\hline
Algorithm & DeepDeblur\cite{nah2017deep}  & SRN-Deblur\cite{tao2018scale}  & PSS-NSC\cite{gao2019dynamic}  & DeblurGANv2\cite{kupyn2019deblurgan} & SVRNN\cite{zhang2018dynamic} & RADN-Deblur \cite{purohit2019region} & GCResNet\\
\hline
PSNR      & 29.08   & 30.26   & 30.92     & 29.55     & 29.19  & 31.76  & \textbf{32.64}   \\
SSIM      & 0.9135  & 0.9432  & 0.9421    & 0.9340    & 0.9306 & 0.9530 & \textbf{0.9580}  \\
FSIM      & 0.9633  & 0.9653  & 0.9756    & 0.9527    & 0.9446 & 0.9798 & \textbf{0.9802}  \\
\hline
\end{tabular}
\end{table*}

\begin{table}[t]
\renewcommand{\arraystretch}{1.3}
\caption{Testing results on HIDE datasets}
\label{result HIDE}
\centering
\begin{tabular}{c|cc|cc}
\hline
\multirow{2}*{Algorithm} & \multicolumn{2}{c|}{HIDE I (long-short)} & \multicolumn{2}{c}{HIDE II (close-ups)} \\
\cline{2-5}
~ & PSNR & SSIM & PSNR & SSIM \\
\hline
DeepDeblur\cite{nah2017deep}           & 27.43 & 0.9020        & 26.18 & 0.8780 \\
SRN-Deblur\cite{tao2018scale}          & 29.41 & 0.9137        & 27.54 & 0.9070 \\
PSS-NSC\cite{gao2019dynamic}           & 29.98 & 0.9234        & 28.14 & 0.9021 \\
DeblurGANv2\cite{kupyn2019deblurgan}   & 28.29 & 0.8960        & 26.64 & 0.8722 \\
RADN-Deblur \cite{purohit2019region}   & 28.97 & 0.9044        & 26.51 & 0.8698 \\
GCResNet & \textbf{30.04} & \textbf{0.9240} & \textbf{28.62} & \textbf{0.9132} \\
\hline
\end{tabular}
\end{table}

\subsubsection{Ablation Study}
\begin{figure}
\centerline{\includegraphics[width=\linewidth]{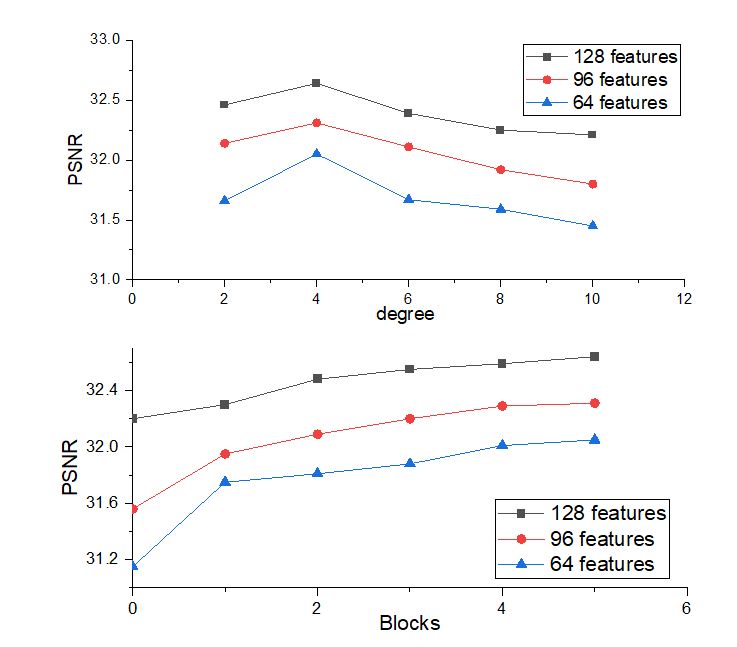}}
\caption{Comparison of different network settings in image deblurring. Degree = 4 achieved the best results with other settings remained the same. Degree = 2 was better than degree = 6, for degree = 6 was too dense so that the advantage of graph convolution could not be produced and hence the poor improvement. Deeper networks showed better performance, but too many residual blocks would not bring further improvement.}
\label{fig:origin}
\end{figure}
We conducted an ablation study on the GCN structure, to show differences between different sets of GC layers. Results are shown in Fig. \ref{fig:origin}. When we kept the number of ResGCN blocks and enlarged the average degree, the performance would decrease for $degree > 4$. Note that the average degree cannot be odd number \cite{watts1998collective}. We think such phenomenon was due to too sparse or too dense connections, and the graph on $degree = 4$ showed the best topological properties. The performance of $degree = 2$ was slightly better than that of $degree = 4$. We imply that the performance would be close to the network without GCN when the average degree kept increasing, because too many connections would make GC meaningless on information. We kept $degree = 4$ and increased the number of ResGCN blocks, the performance kept increasing while the growth rate decreased with too many blocks. The network without ResGCN had the worst performance. In addition, performance could be improved by using more features, but that would also lead to heavier network in CNN. Consider the limit of memory, network with 128 features and degree of 4 showed the best balanced performance. We used 5 ResGCN blocks in the GCResNet.


\subsection{Experiments on Super-resolution}
\subsubsection{GCEDSR Network}
The network for super-resolution consists of graph convolutions with the EDSR \cite{lim2017enhanced} and we term it as Graph Convolution EDSR (GCEDSR). We adopted graph convolution after 8 Resblocks, and then followed by another 8 Resblocks until the final SR image produced. We used 5 ResGCN blocks and $degree = 2$, based on the ablation study on deblurring. For a fair comparison, we did not change any other settings in the EDSR, except for adding graph convolutions. All the convolution layers had 256 channels. 

We used L1 loss as the loss function, as it is the most suitable loss function and widely used in single image super-resolution \cite{haris2018deep, lim2017enhanced}. 

\begin{table*}[t]
\begin{center}
\caption{Quantitative results with BI degradation model. Best results are highlighted.}
\label{table:SR}
\begin{tabular}{l|c|c|c|c|c}
\hline
Methods & Scale & Set5 & Set14 & Urban100  & B100 \\

\hline
SRCNN \cite{dong2015image} & $\times2$ & 36.66 / 0.9542 & 32.45 / 0.9067 & 29.50 / 0.8946 & 31.36 / 0.8879 \\
FSRCNN \cite{dong2016accelerating} & $\times2$ & 37.05 / 0.9560 & 32.66 / 0.9090 & 29.88 / 0.9020 & 31.53 / 0.8920 \\
VDSR \cite{kim2016accurate} & $\times2$ & 37.53 / 0.9590 & 33.05 / 0.9130 & 30.77 / 0.9140 & 31.90 / 0.8960 \\
RDN \cite{zhang2018residual} & $\times2$ & 38.24 / 0.9614 & 34.01 / 0.9212 & 32.89 / 0.9353 & 32.34 / 0.9017 \\
D-DBPN \cite{haris2018deep} & $\times2$ & 38.09 / 0.9600 & 33.85 / 0.9190 & 32.55 / 0.9324 & 32.27 / 0.9000 \\
EDSR \cite{lim2017enhanced} & $\times2$ & 38.11 / 0.9602 & 33.92 / 0.9195 & 32.93 / 0.9351 & 32.32 / 0.9013 \\
GCEDSR & $\times2$ & \textbf{38.29} / \textbf{0.9615} & \textbf{34.05} / \textbf{0.9213} & \textbf{33.12} / \textbf{0.9386} & \textbf{32.39} / \textbf{0.9023} \\
\hline
SRCNN \cite{dong2015image}& $\times4$ & 30.48 / 0.8628 & 27.50 / 0.7513 & 24.52 / 0.7221 & 26.90 / 0.7101 \\
FSRCNN \cite{dong2016accelerating} & $\times4$ & 30.72 / 0.8660 & 27.61 / 0.7550 & 24.62 / 0.7280 & 26.98 / 0.7150 \\
VDSR \cite{kim2016accurate} & $\times4$ & 31.35 / 0.8830 & 28.02 / 0.7680 & 25.18 / 0.7540 & 27.29 / 0.7260 \\
RDN \cite{zhang2018residual} & $\times4$ & 32.47 / 0.8990 & 28.81 / 0.7871 & 26.61 / 0.8028 & 27.72 / 0.7419 \\
D-DBPN \cite{haris2018deep} & $\times4$ & 32.47 / 0.8980 & 28.82 / 0.7860 & 26.38 / 0.7946 & 27.72 / 0.7400 \\
EDSR \cite{lim2017enhanced} & $\times4$ & 32.46 / 0.8968 & 28.80 / 0.7876 & 26.64 / 0.8033 & 27.71 / 0.7420 \\
GCEDSR & $\times4$ & \textbf{32.61} / \textbf{0.9001} & \textbf{28.89} / \textbf{0.7885} & \textbf{26.72} / \textbf{0.8079} & \textbf{27.76} / \textbf{0.7439} \\
\hline
SRCNN \cite{dong2015image}& $\times8$ & 25.33 / 0.6900 & 23.76 / 0.5910 & 21.29 / 0.5440 & 24.13 / 0.5660  \\
FSRCNN \cite{dong2016accelerating} & $\times8$ & 20.13 / 0.5520 & 19.75 / 0.4820 & 21.32 / 0.5380 & 24.21 / 0.5680 \\
VDSR \cite{kim2016accurate} & $\times8$ & 25.93 / 0.7240 & 24.26 / 0.6140 & 21.70 / 0.5710 & 24.49 / 0.5830 \\
D-DBPN \cite{haris2018deep} & $\times8$ & 27.21 / 0.7840 & 25.13 / 0.6480 & 22.73 / 0.6312 & 24.88 / 0.6010 \\
EDSR \cite{lim2017enhanced} & $\times8$ & 26.96 / 0.7762 & 24.91 / 0.6420 & 22.51 / 0.6221 & 24.81 / 0.5985 \\
GCEDSR & $\times8$ & \textbf{27.39} / \textbf{0.7876} & \textbf{25.18} / \textbf{0.6503} & \textbf{23.14} / \textbf{0.6370} & \textbf{24.92} / \textbf{0.6027} \\
\hline
\end{tabular}
\end{center}
\end{table*}
\subsubsection{Training Datasets}
Following \cite{lim2017enhanced}, \cite{zhang2018image}, we used 800 training images from DIV2K dataset \cite{timofte2017ntire} as training set. We used four standard benchmark datasets for testing: Set5 \cite{bevilacqua2012low}, Set14 \cite{zeyde2010single}, B100 \cite{martin2001database}, and Urban100 \cite{huang2015single}. We conducted experiments with Bicubic (BI) . The SR results were evaluated with PSNR and SSIM. 
\subsubsection{Training Details}
We implemented the proposed network by Pytorch on a NVIDIA Tesla P100 GPU. During training, we randomly cropped a $48 \times 48$ region from a blurred image as training input, along with its ground truth image at the same location. The batch size was set to 24. All weights were initialized by the Xavier method \cite{glorot2010understanding}, and biases were initialized to zero. The network was optimized by using the Adam \cite{kingma2014adam} with default setting $\beta_1=0.9$, $\beta_2=0.999$ and $\epsilon=10^{-8}$.
\subsubsection{Results}
Quantitative results are demonstrated in Table \ref{table:SR} and qualitative results in Fig. \ref{fig:srres}. One can observe from Fig. \ref{fig:srres} that the proposed network outperforms the previous methods at $4 \times$ super-resolution levels. 

\section{Analysis}
\subsection{With Laplacian Regularization}
In the proposed GCN, feature maps of the encoder are first placed onto a graph structure. Then in the GC layers, these features, now nodes, further undergo graph convolutions, by Eq. (\ref{eq:GC}). These GC layers inexplicitly apply graph Laplacian regularization \cite{kipf2016semi}, to the resulting the feature maps, $X \Theta$, of the encoder. The proposed approach combines the efficacy of CNNs in feature extraction and effectiveness of GNNs for constraining feature relationships. With a predefined graph structure, the method is also extremely efficient.

\subsection{With Channel Attention}
The proposed method can be regarded as an expansion of channel attention mechanism. Consider a simple example: for a network that only connects all nodes to 8 special nodes respectively. Hence the graph convolution in our method is similar with channel-wise attention focused on these 8 channels, which has a similar structure in \cite{zhang2018image}. The graph that we use is small-world graph, with small average path length, thus key nodes can extract information from other vertexes within few GC layers.

\section{Conclusion}
In this paper, we proposed a new convolutional neural network for image deblurring and super-resolution by adapting graph network in CNNs. While existing graph neural networks are for image classification, the proposed network explore graph structures in the feature maps of CNNs for effective image restoration. Experiments demonstrate that such adaptation with a predefined graph structure can achieve improved performance in image restoration with little added computational costs. Exploring topological relationships among feature maps is beneficial to many image processing tasks.

\bibliographystyle{IEEEtran}
\bibliography{main}

\end{document}